# Deep Generative Design:
# Integration of Topology Optimization and Generative Models

Sangeun Oh[1,†], Yongsu Jung[2,†], Seongsin Kim[1], Ikjin Lee[2,*], Namwoo Kang[1,*]

[1]Sookmyung Women's University, Cheongpa-ro 47-gil 100, Yongsan-gu, Seoul, Korea

[2]Korea Advanced Institute of Science and Technology, 291, Daehak-ro, Yuseong-gu, Daejeon, Korea

[†]S. Oh and Y. Jung contributed equally to this work.

[*]Co-corresponding authors: ikjin.lee@kaist.ac.kr (I. Lee), nwkang@sm.ac.kr (N. Kang)

**Abstract**

Deep learning has recently been applied to various research areas of design optimization. This study presents the need and effectiveness of adopting deep learning for generative design (or design exploration) research area. This work proposes an artificial intelligent (AI)-based deep generative design framework that is capable of generating numerous design options which are not only aesthetic but also optimized for engineering performance. The proposed framework integrates topology optimization and generative models (e.g., generative adversarial networks (GANs)) in an iterative manner to explore new design options, thus generating a large number of designs starting from limited previous design data. In addition, anomaly detection can evaluate the novelty of generated designs, thus helping designers choose among design options. The 2D wheel design problem is applied as a case study for validation of the proposed framework. The framework manifests better aesthetics, diversity, and robustness of generated designs than previous generative design methods.

## Nomenclature

$\mathbb{E}$: expectation
$D$: differentiable function of discriminator
$G$: differentiable function of generator
$z$: noise variable
$\mathcal{L}$: loss function of autoencoder
c: compliance
$x_e$: density variable
$\tilde{x}_e$: filtered density variable
$\bar{\tilde{x}}_e$: projected density variable
$p$: penalization factor

# 1. Introduction

Artificial intelligence (AI) covers all technologies pursuing machines to imitate human behavior. Machine learning is a subset of AI, which attempts to learn meaningful patterns from raw data by using statistical methods. Deep learning seeks to enhance the learning ability with a hierarchical neural network structure that consists of several layers [1,2]. Recently, deep learning has been employed not only in computer science but also in various engineering domains. A physics-based approach in engineering domains can be replaced with a data-driven approach in an effective way. In mechanical engineering, it has been widely applied to autonomous driving, robot control, biomedical engineering, prognostics and health management (PHM), and design optimization.

Deep learning research related to design optimization can be classified as follows: (1) topology optimization [3-7], (2) shape parameterization [8,9], (3) computer-aided engineering (CAE) simulation and meta modeling [10-12], (4) material design [13-15], and (5) design preference estimation [16,17]. Section 2 introduces each research in detail.

This study commenced from the idea that deep learning is indispensable for generative design (i.e., design exploration), and the aforementioned deep learning research can certainly be integrated through the proposed framework in a holistic view.

## 1.1. Generative Design Using Topology Optimization

Generative design is one of the design exploration methods performed by typically varying design geometry parametrically and assessing the performance of output designs [18,19]. Recent research on generative design utilizes topology optimization as a design generator instead of design parameterization and develop the methods to generate numerous designs in parallel with cloud computing [20]. A designer provides diverse boundary conditions of topology optimization, which brings different optimized designs under different boundary conditions. Matejka et al. [21] state "generative design varies the parameters of the problem definition while parametric design varies parameters of the geometry directly."

Generative design aspires to explore the design options that satisfy structural performance and choose suitable designs for various designers' needs, whereas conventional topology optimization seeks to find an optimal design. The generative design concept has been rapidly developed and implemented in a commercial software [22] and applied in designing various structures such as automobile, architecture, and aircraft.

The overall process of generative design consists of four stages as follows [20]:
- Step 1: Set the design parameters and goals for topology optimization.
- Step 2: Generate designs with running topology optimization under different parameters.
- Step 3: Study options, iterate, and select the best design.
- Step 4: Manufacture the design by 3D printing.

In particular, 3D printing technology development enabled the production of complicated geometric designs, which further accelerates the practical use of generative design.

However, several drawbacks in current generative design were identified [23]. First, it does not use state-of-the-art AI technology (e.g., deep learning), even though topology optimization can be considered as AI in a broad way. Second, it is unable to create aesthetic designs. Topology optimization focuses solely on engineering performance. Therefore, the results seem to be counterintuitive in the aesthetic point of view. However, aesthetics is an essential factor for customers and should be balanced with engineering performance [24]. Third, the diversity of optimized designs is low, which can result in a new design in terms of intensity or density of pixels. However, these designs may be similar in terms of human perception.

## 1.2. Generative Models for Generative Design

Generative models, one of the promising deep learning areas, can enhance research on generative design. The generative model is an algorithm for constructing a generator that learns the probability distribution of training data and generates new data based on learned probability distribution. In particular, variational autoencoder (VAE) and generative adversarial network (GAN) are popular generative models used in design optimization, where high-dimensional design variables are encoded in low-dimensional design space [13,14]. In addition, these models are utilized in the design exploration and shape parameterization [8,9].

The use of generative model to produce engineering designs directly is limited [23]. However, this study claims that the limitations can be overcome by integrating with topology optimization. First, the generative model requires a number of training data, but accumulated training data for various designs in the industry are confidential and difficult to access. A number of designs obtained from topology optimization are expected to serve as training data. Second, the generative model cannot guarantee feasible engineering. In this case, engineering performance can be evaluated through topology optimization. Third, mode collapse is one of the main problems in the generative model, producing only specific results and bringing large variance of output quality. However, low-quality designs can be improved through post-processing by employing topology optimization.

### 1.3. Research Purpose

This study proposes the new framework for generative design by integrating topology optimization and generative models. The proposed framework can provide a number of meaningful design options accounting for engineering performance and aesthetics and allows evaluation and visualization of the new design options according to the design attributes (e.g., novelty, compliance, and cost).

The proposed framework consists of iterative design exploration and design evaluation parts. Iterative design exploration involves generating a large number of new designs iteratively by using a small number of previous designs. Design evaluation involves quantifying the novelty of generated designs in comparison with the previous designs and visualizing design options with other design attributes. The proposed framework is applied to 2D wheel design of an automobile for demonstration.

The rest of this study is structured as follows. Section 2 reviews previous design optimization studies that employ deep learning. Section 3 proposes a deep generative design framework. Sections 4 and 5 present topology optimization and generative models, respectively, which are the main methodologies used in our study. Sections 6 and 7 present and discuss case study results, respectively. Finally, Section 8 summarizes the conclusions and limitation and introduces future work.

## 2. Literature Review: Deep Learning in Design Optimization Research

The deep learning-based design optimization research is explained as follows. First, topology optimization can be interpreted as deep learning from the perspective of pixel-wise image labeling because it distributes the materials in design domain accounting for objective and constraint functions [3]. Intensive computational demand is a drawback of topology optimization due to iterative finite element analysis (FEA) of a structure. Thus, Yu et al. [4] propose the framework where a low-resolution structure is first generated by using a convolutional neural network (CNN)-based encoder and decoder constructed on 2D topology optimal designs, and then it is upscaled to a high-resolution structure through conditional GAN (cGAN) without any iterative FEA in topology optimization. Banga et al. [5] employ 3D CNN-based encoder and decoder, which allows the final optimized structural output to be obtained from intermediate structural inputs. Guo et al. [6] perform the topology optimization on latent variables of reduced dimensional space by using VAE and style transfer. Cang et al. [7] used active learning to constrain the training of a neural network so that the network results in near-optimal topologies.

Second, deep learning applications on shape parameterization (i.e., design representation) have been developed. The parameterization to define the geometry is necessary for shape optimization. However, defining the variables in complicated geometry is extremely difficult, and the correlation between variables is too strong, which can hinder parameterization in a mathematical approach. Burnap et al. [8] show the parameterization possibility of the 2D shape of an automobile by using VAE. Moreover, Umetani [9] shows the parameterization of 3D meshes of an automobile by using autoencoder.

Third, deep learning has been applied to the metamodeling and simulation-based optimization. Many researches tried to apply deep learning to computational fluid dynamics (CFD) because CFD simulation has high computational cost. Guo et al. [10] propose the CNN model to predict the responses of CFD simulation for the 2D shape of an automobile, and Tompson et al. [11] accelerate Eulerian fluid simulation by approximating linear system with CNN. Farimani et al. [12] also propose the integration of cGAN to solve the steady-state problem for heat conduction and incompressible fluid flow.

Fourth, deep learning application on material design has been developed because of the direct relationship between the density of element in material structure and pixels of images, resulting to easy transformation of the domain from the material structures to images. Yang et al. [14] obtain optimal microstructure by using Bayesian optimization framework where the microstructure is mapped into low-dimensional latent variables by using GAN. Cang et al. [13] propose a feature extraction method to convert the microstructure to low-dimensional design space through a convolutional deep belief network. Cang et al. [15] show that generating an arbitrary amount of microstructure by a small amount of training data is available, proposing VAE under morphology constraint.

Finally, the application of deep learning on design preference estimation has been developed. Burnap et al. [16] improve the prediction accuracy of the customer's preference model by learning restricted Boltzmann machine (RBM) with the original design variables as input and extracting the future. Pan et al. [17] propose the learning preference on aesthetic appeal using Siamese neural network architecture with cGAN.

In addition, deep learning is used in various engineering designs. For instance, Dering and Tucker [25] have successfully mapped the form and function of the design by using 3D convolution neural network, and Dering and Tucker [26] propose an image generation model that integrates deep learning and big data.

The categorization of the aforementioned research is not independent from one another, thus allowing integration and subsequently enhancing the conventional design optimization process. Especially, the authors claim that generative design is located at the intersection of all these research areas, and that it would be a very promising research area within an AI-based design automation system.

## 3. Deep Generative Design Framework

A deep generative design framework was proposed, which integrates topology optimization and generative models. Fig. 1 shows the entire process which consists of two main parts (i.e., iterative design exploration and design evaluation) and nine stages.

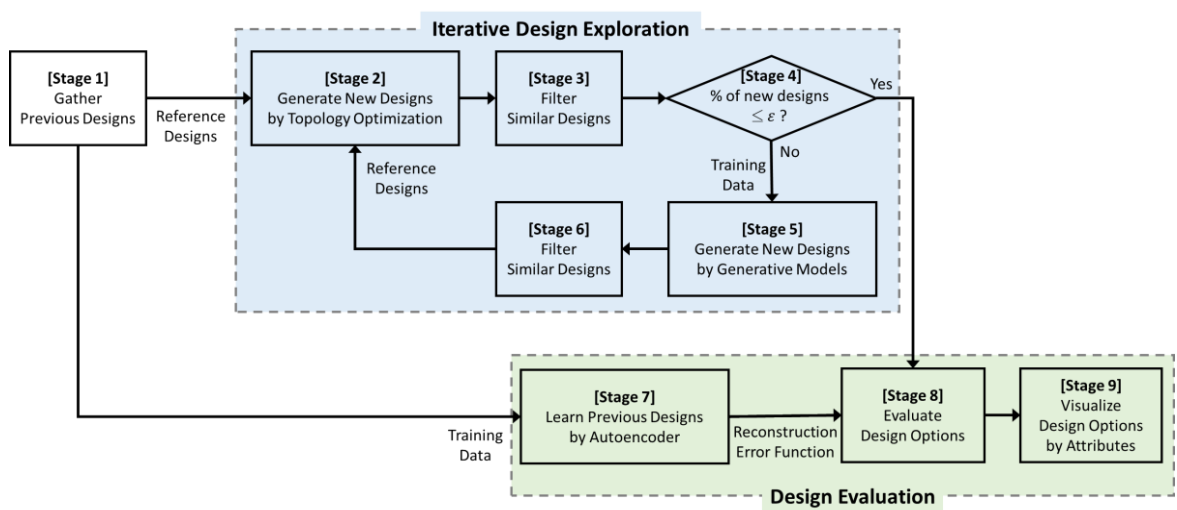

**Fig. 1.** Deep generative design framework

The iterative design exploration is the integration of topology optimization and generative models to produce new designs, and design evaluation quantifies and evaluates the novelty and main attributes of new designs. Each stage is explained as follows:

In Stage 1, the previous designs in the market and the industry are collected as reference designs for Stage 2. In this study, the reference design is defined as a benchmark design to create new designs in topology optimization.

In Stage 2, new designs are obtained by topology optimization based on reference designs. In this step, topology optimization has a multi-objective function of (1) compliance minimization, which represents engineering performance, and (2) difference (i.e., pixel-wise L1 distance) minimization from the reference design, which aims to improve aesthetics and diversity. Different designs are obtained by varying the relative weights of each objective function because a tradeoff between two objectives exists. Here, we assume that previous designs in the market are more aesthetic than conventional topology optimization results, because previous designs are created by human designers. If topology optimization can benchmark the shape of previous designs, the final optimization result is expected to be more aesthetic. In terms of diversity, the more diverse reference designs are used as input, the more diverse topology optimization results would be. Detail of this stage is explained in Section 4.2.

In Stage 3, similar designs gathered from topology optimization are filtered out by the similarity criterion configured for the user-specified threshold to reduce computational costs due to irrelevant designs. This study uses the pixel-wise L1 distance as a criterion to judge a similar design, which is set to $10^3$. If we set this value much tightly, then the number of designs to be generated will be reduced, but the differentiation between designs will be improved. This process is repeated in Stage 6 to filter out the designs generated from the generative models. It is important to note that L1-norm distinguishes two of the same designs with different rotations, and considers them to be different. In order to resolve this limitation of L1-norm, one can map the design variables into the latent space through generative models, and calculate the L1-norm. In our case study, however, there were not many of these cases, so we used L1-norm in the design space only.

In Stage 4, the ratio of the number of new designs in the current iteration to the number of total designs in the previous iteration is calculated. If it is smaller than the user-specified threshold, then exit the iterative design exploration and jump to Stage 8; otherwise, proceed to Stage 5. In our study, the threshold for termination criteria is set to 0.3. This value can be adjusted according to how much diverse designs the user wants to generate.

In Stage 5, new designs are created by generative models after learning aggregated designs in the current iteration, and they are used as reference designs in Stage 2 after filtering out similar designs in Stage 6. We used Boundary Equilibrium GAN (BEGAN), whose structure and settings are introduced in Section 5.2. The iterative design exploration should be continuously performed from Stage 2 to 6 until the termination criterion is satisfied, i.e., until the amount of generated designs are substantial. This iterative process has the purpose of creating a large number of various designs by starting from a small number of previous designs in Stage 1.

Next, design evaluation part consists of Stages 7 to 9. Stage 7 involves the building of a loss function (i.e., reconstruction error function) employing autoencoder trained by previous designs of Stage 1. This function can be used to evaluate design novelty in comparison with the previous designs. The details of autoencoder model are introduced in Section 5.2. In Stage 8, design options obtained from iterative design exploration have to be evaluated on the basis of various design attributes that are essential to the designers. It can evaluate not only the novelty of generated designs but also the physical quantities such as the volume and compliance of designs. Finally, the tradeoff between attributes is demonstrated as plotting of designs in each axis of design attributes, and the proper designs can be chosen accounting to the relative importance of each attribute.

The proposed framework is applied to the 2D wheel of an automobile as a case study. The following Sections 4 and 5 present detailed descriptions of two main methodologies of our proposed framework, i.e. topology optimization and generative models.

## 4. Topology Optimization

Section 4.1 introduces the basic theory of topology optimization which our study stems from, and Section 4.2 presents the proposed topology optimization method for the wheel design case study.

### 4.1 Basic Theory

#### 4.1.1. Density-based Approach

Topology optimization is commonly referred to the material distribution method developed and spread to a wide range of disciplines. The basic concept is how to distribute materials in a given design domain without any preconceived design [27-29]. In this study, the compliance minimization related to the stiffness of a structure has been carried out to redesign existing wheels. Many approaches such as homogenization and level-set methods can be applied, but we choose the density-based approach where material distribution is parameterized by the density of elements. Especially, solid isotropic material with penalization (SIMP) is a procedure that implicitly penalizes intermediate density values to lead to the black-and-white design. The basic formulation of SIMP in compliance minimization can be written as [29,30]

$$\begin{aligned} \min \quad & c(\mathbf{x}) = \mathbf{U}^T \mathbf{K} \mathbf{U} = \sum_{e=1}^{N_e} \mathbf{u}_e^T \left( E_e(x_e) \mathbf{k}_0 \right) \mathbf{u}_e \\ \text{s.t} \quad & V(\mathbf{x})/V_0 = f \\ & \mathbf{K}\mathbf{U} = \mathbf{F} \\ & 0 \le x_e \le 1, \ e = 1,...,N_e \end{aligned} \quad (1)$$

where $\mathbf{U}$ is a displacement vector; $\mathbf{K}$ is a global stiffness matrix; $c(\mathbf{x})$ is the compliance; $\mathbf{u}_e$ is an element displacement vector; $\mathbf{k}_0$ is an element stiffness matrix; $f$ is the volume fraction; $N_e$ is the number of elements; $x_e$ is the design variable (i.e., density) of element $e$; and $V(\mathbf{x})$ and $V_0$ are the material volume and the volume of design domain, respectively. In modified SIMP, the density that is directly associated with Young's modulus can be expressed as [31]

$$E_e(x_e) = E_{\min} + x_e^p (E_0 - E_{\min}) \quad (2)$$

where $p$ is a penalization factor to ensure the black-and-white design, and $E_{\min}$ is introduced to avoid numerical instability when the density of elements become zero.

Many studies have been done to enhance the performance of topology optimization such as filtering techniques. In this study, we develop the code based on 99- and 88-line MATLAB codes, which are the simplest and most efficient two-dimensional topology optimization codes written in MATLAB [30,32]. Thus, we will briefly explain algorithms such as sensitivity analyses and filtering techniques used in this study.

**4.1.2 Sensitivity Analysis and Filtering Techniques**

In gradient-based optimization, the sensitivity analysis of objective and constraint function with respect to each design variable is required to provide accurate search direction to the optimizer. Therefore, the sensitivity analysis with respect to the density of elements can be given by

$$\frac{\partial c}{\partial x_e} = -p x_e^{p-1}(E_0 - E_{\min})\mathbf{u}_e^T \mathbf{k}_0 \mathbf{u}_e \qquad (3)$$

and

$$\frac{\partial V}{\partial x_e} = \frac{\partial}{\partial x_e}\left(\sum_{e=1}^{N_e} x_e v_e\right) = 1 \qquad (4)$$

under the assumption that all elements have a unit volume. On the other hand, the optimality criteria (OC) method, one of the classical approaches to structural optimization problems, is employed in this paper. The OC method updates the design domain as

$$x_e^{new} = \begin{cases} \max(0, x_e - m) & \text{if } x_e B_e^\eta \leq \max(0, x_e - m) \\ \min(1, x_e + m) & \text{if } x_e B_e^\eta \geq \min(1, x_e + m) \\ x_e B_e^\eta & \text{otherwise} \end{cases} \qquad (5)$$

where $m$ is a positive move-limit and $\eta$ is a numerical damping coefficient, and

$$B_e = \frac{-\dfrac{\partial c}{\partial x_e}}{\lambda \dfrac{\partial V}{\partial x_e}} \qquad (6)$$

The Lagrange multiplier related to volume fraction constraint can be obtained from a bisection algorithm that is one of the popular root-finding algorithms. The termination criteria for the convergence can be written as

$$\|\mathbf{x}_{new} - \mathbf{x}\|_\infty \leq \varepsilon \qquad (7)$$

where $\varepsilon$ is the tolerance usually set as a relatively small value such as 0.01.

For the assurance of the existence of well-posed and mesh-independent solutions, several strategies to avoid a checkerboard pattern and gray-scale issues are introduced. In this study, we apply so-called three-field SIMP, which has a projection scheme. Three-field means the original density, filtered density, and projected density. Detailed descriptions can be seen in the literature [33].

The basic filters applied to topology optimization are sensitivity and density filters, which are used in one-field and two-field SIMP, respectively. The main idea of both techniques is to modify sensitivity or physical element density to be a weighted average of the neighborhood. The neighborhood is defined on the basis of the distance from the center of the element, and the maximum distance to include in the neighborhood is a user-specified parameter referred to the mesh-independent radius. The sensitivity filter can be written as [34]

$$\frac{\partial c}{\partial x_e} = \frac{1}{x_e \sum_{f=1}^{N} H_f} \sum_{f=1}^{N} H_f x_f \frac{\partial c}{\partial x_f} \tag{8}$$

where the convolution operator can be written as

$$H_f = r_{\min} - \text{dist}(e, f), \tag{9}$$

where subscript $f$ means one of the elements that the center-to-center distance expressed as $\text{dist}(e, f)$ between elements is smaller than $r_{\min}$.

The density filter defines the physical density with weighted averaging. The weighted average concept is the same in the sensitivity filter as Eq. (8), but the density is filtered instead of sensitivity expressed as [35,36]

$$\tilde{x}_e = \frac{\sum_{f=1}^{N} H_f x_f}{\sum_{f=1}^{N} H_f} \tag{10}$$

Therefore, original and filtered densities can be referred to as a design variable and physical density, respectively. The sensitivity analysis with respect to design variables should be modified by introducing the physical density using a chain rule. A detailed description can be seen in the literature [37].

The weighted average is used in both filtering methods to avoid the checker-board pattern in an optimum design. However, the density filter can induce gray transitions between solid and void regions. Thus, the third field of density, or the so-called projection filter, is introduced. It mitigates the gray transition problem by projecting to solid and void usually using a smoothed Heaviside projection [31,37,38]. In this study, we use the Heaviside projection filter on the filtered density obtained from Eq. (10). The projection filter can be written as

$$\bar{\tilde{x}}_e = 1 - e^{-\beta \tilde{x}_e} + \tilde{x}_e e^{-\beta} \tag{11}$$

where $\beta$ is a parameter related to slope of projection and can be updated through the optimization. In three-field SIMP with projection filter, the sensitivity analysis is modified compared with Eq. (8) because the finite element analysis is performed based on the physical density obtained from Eq. (11). The sensitivity analysis with respect to design variables can be easily derived using the chain rule.

### 4.2. Proposed Topology Optimization

In the deep generative design framework of Fig. 1, Stage 2 generates new engineering designs through topology optimization reflecting the shape of the reference designs that can be either previous wheel designs from Stage 1 or generated designs from generative models (Stages 5 and 6). A number of engineering performances has to be considered when designing the wheel of vehicle, but compliance obtained from the static analysis has been generally employed in this research for the sake of simplicity. Fig. 2 sketches the design domain and boundary conditions for 2D wheel design. The original design domain is 128 by 128 elements, and the reference designs also have 128 by 128 pixels. The outer ring of the wheel is set to the non-design region to maintain the shape of the rim, and the inner region is set to fixed boundary condition for connecting parts. Therefore, spoke is the main component in the design domain.

The element stiffness matrix is based on 4-node bilinear square elements in the 88-line MATLAB code [32]. Normal and shear forces are uniformly exerted along the surface, which are common load conditions in the 2D wheel optimization. The normal force and the shear force represent uniform tire pressure and tangential traction, respectively. The ground reaction induced by vehicle weight is disregarded because it requires an additional symmetric condition.

The ratio between normal and shear force is a user-specified parameter that can significantly change the optimized wheel design. The force ratio is defined as the magnitude of normal force divided by shear force.

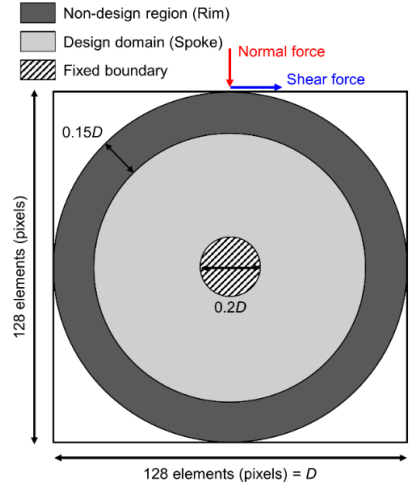

**Fig. 2.** Design domain and boundary conditions of a 2D wheel design

However, varying boundary conditions only can limit the production of meaningful and diverse designs. Thus, a new objective function in topology optimization was introduced so that the design generator can produce engineering designs while maintaining the shape of various reference designs. The modified objective function can be formulated as

$$f(\mathbf{x}) = \mathbf{U}^T \mathbf{K}(\mathbf{x}) \mathbf{U} + \lambda \left\| \mathbf{x}^* - \mathbf{x} \right\|_1 \tag{12}$$

where $\mathbf{U}^T \mathbf{K}(\mathbf{x}) \mathbf{U}$ is compliance, $\lambda$ is a user-specified similarity parameter, and $\mathbf{x}^*$ is elements of the reference design. Therefore, the L1-norm between generated and reference design represents the similarity. L1-norm is more preferred than L2-norm because it can alleviate the blurring of design, and shows better design quality. In addition, cross entropy loss can also be an alternative to L1-norm depending on how the problem is defined. However, since cross entropy is more commonly used in discrete problems, the pixel-wise L1-norm is more appropriate in our problem which is continuous.

All other processes are identical with conventional three-field SIMP explained in Section 4.1. Reference design represented as $\mathbf{x}^*$ is a binary matrix with entries from the Boolean domain since it is black-and-white design. Hence, sensitivity analysis for additional similarity term can be expressed as

$$\frac{\partial}{\partial \mathbf{x}} \left( \lambda \left\| \mathbf{x}^* - \mathbf{x} \right\|_1 \right) \cong -\lambda \mathbf{x}^* \tag{13}$$

The above expression means that if a specific element in reference design is solid, then the sensitivity is set to $-\lambda$, and 0 otherwise to avoid providing the positive sensitivity to the OC optimizer. In other words, the purpose of Eq. (13) is to give additional weights of sensitivity on the solid elements of reference design, so that the optimized design can be affected by the shape of reference design.

Consequently, five discrete levels of similarity parameter and force ratio are configured to generate new designs from topology optimization as listed in Table 1. For instance, if 100 reference designs are available, then 2500 designs (100 reference designs × 5 similarity parameter levels × 5 force ratio levels) can be obtained from topology optimization. The type of condition and the number of levels can be determined by the designers such as voxel sizes, solver parameters, and the number of iterations in topology optimization [21].

**Table 1.** Condition parameter levels

| Condition | Levels |
|---|---|
| Similarity | 5 levels: 0.0005/0.005/0.05/0.5/5 |
| Force ratio | 5 levels: 0/0.1/0.2/0.3/0.4 |

## 5. Generative Models

Section 5.1 introduces popular generative models briefly, and explains the BEGAN model which is mainly used in our proposed framework. Section 5.2 shows how to utilize BEGAN for generating and evaluating wheel designs.

## 5.1. Basic Theory

### 5.1.1. Generative Adversarial Networks (GANs)

GANs are designed to infer the data generating a distribution $p_{\text{data}}(x)$ by making the model distribution generated by generator $p_{\text{g}}(G)$ to be close to the real data distribution where $x$ is the variable of real data and $G$ is the generator's differentiable function with parameters $\theta_{\text{g}}$. Function $G$ has an input noise variable $z$ and tries to map it to the real data space by adjusting $\theta_{\text{g}}$, thus represented as $G(z; \theta_{\text{g}})$. Similarly, the discriminator's differentiable function is derived as $D(x; \theta_{\text{d}})$, which attempts to predict the probability that the input is from the real dataset. The zero-sum game of maximizing the discriminator and generator is equivalent to maximizing $\log D(x)$ and minimizing $\log(1- D(G(z)))$ [39,40] as

$$\min_G \max_D V(D, G) = \mathbb{E}_{x \sim p_{data}(x)}[\log D(x)] + \mathbb{E}_{z \sim p_z(z)}[\log(1- D(G(z))] \quad (14)$$

With this standard GAN structure, various GANs have been developed by modifying the generator, discriminator, or objective function.

To overcome the notorious difficulty in training GANs, deep convolutional GANs (DCGANs) provide a stable training model, which works on various datasets by constructing a convolutional neural network in the generator and discriminator. In addition, DCGANs suggest certain techniques such as removing a fully-connected layer on top, applying batch normalization, and using the leaky rectified linear unit activation function [41].

Adversarially learned inference (ALI) and bidirectional GANs (BiGANs) adopt the encoding–decoding model to the generator to improve the quality of generated samples in an efficient way. Moreover, BiGANs emphasize taking advantages of learned features [42,43]. For high-resolution images with stable convergence and scalability, energy-based GANs (EBGANs) have proven to produce realistic 128 × 128 images. EBGANs consider the discriminator as an energy function and the energy as the reconstruction error. From this point of view, an autoencoder architecture is used for the discriminator [44]. The autoencoder consists of encoder and decoder functions. The input value is transformed through the encoder and is restored to its original form again through the decoder [2]. Wasserstein GANs (WGANs) approach the way to obtain good image quality by changing the distance measure of two probability distributions. WGANs show that the earth-mover distance, which is also called Wasserstein-1, provides a differentiable function, and, thus, produces meaningful gradients, whereas Kullback-Leibler and Jensen-Shannon divergence in previous research do not when two probability distributions are disjointed [40]. BEGANs also use the Wasserstein distance as a measure of convergence. BEGANs present an equilibrium concept, balancing the discriminator and generator in training and the numerical way of global convergence [45].

Aside from enhancing the image quality, the way to control the mode of generated outputs is presented by cGANs [46] and InfoGAN (Information maximizing Generative Adversarial Nets) [47]. cGANs provide additional input values to the generator and discriminator for categorical image generation. Furthermore, InfoGAN lets the generator produce uncategorical images by adding a latent code that can be categorical and continuous. It is useful for finding hidden representations from large amounts of data. However, intentionally creating a specific image is still difficult.

Hitherto, many studies on GANs contribute to good image quality in terms of convergence and stability. However, GANs are still weak in utilizing from the design engineering point of view, such as uneven image quality from the same saving point of the model, especially when relatively small amounts of training data are given and images have insufficient engineering features.

### 5.1.2. Boundary Equilibrium GAN (BEGAN)

This paper employs BEGAN among GANs for the proposed framework because it provides a robust visual quality in a fast and stable manner. The autoencoder architecture as the discriminator used in EBGANs is also introduced by BEGANs. Similar to WGANs, BEGANs use the Wasserstein distance as a measure of convergence. With these techniques, BEGANs achieve reliable gradients that are difficult for high-resolution 128 × 128 images.

Rather than trying to match the probability distribution of real data, BEGANs focus on matching autoencoder loss distribution. It measures the loss, which is the difference between the sample and its output that passed through the autoencoder. Subsequently, a lower bound of the Wasserstein distance between the autoencoder loss distribution of real and that of generated samples is derived. The autoencoder loss function, $\mathcal{L}: \mathbb{R}^{N_x} \to \mathbb{R}^+$, is defined as

$$\mathcal{L}(v) = |v - A(v)|^\eta$$

where (15)

$$\begin{cases} \mathbb{R}^{N_x} \to \mathbb{R}^{N_x} & \text{is an autoencoder function} \\ \eta \in \{1, 2\} & \text{is a targe norm} \\ v \in \mathbb{R}^{N_z} & \text{is a sample of dimension } N_z \end{cases}$$

Applying Jensens' inequality, the lower bound of the Wasserstein distance is derived as

$$|m_1 - m_2| \qquad (16)$$

where $m_i \in \mathbb{R}$ is the mean of autoencoder loss distribution. For the maximization of Eq. (16) for the discriminator with $m_1 \to 0$ and $m_2 \to \infty$, BEGANs' objective function is described as minimizing the discriminator's autoencoder loss function $\mathcal{L}_D$ and generator's one $\mathcal{L}_G$ as the following where $\theta_D$ and $\theta_G$ are the parameters of the discriminator and generator, $G: \mathbb{R}^{N_z} \to \mathbb{R}^{N_z}$ is the generator function, $z \in [-1,1]^{N_z}$ are uniform random samples of dimension $N_z$, and $z_D$ and $z_G$ are samples from z. The objective functions are defined as

$$\begin{cases} \mathcal{L}_D = \mathcal{L}(x) - k_t \mathcal{L}(G(z_D)) & \text{for } \theta_D \\ \mathcal{L}_G = \mathcal{L}(G(z_G)) & \text{for } \theta_G \\ k_{t+1} = k_t + \lambda_k (\gamma \mathcal{L}(x) - \mathcal{L}(G(z_G))) & \text{for } \gamma = \frac{\mathbb{E}[\mathcal{L}(G(z))]}{\mathbb{E}[\mathcal{L}(x)]} \end{cases} \qquad (17)$$

where $k_t \in [0,1]$ is a control factor to determine how much $\mathcal{L}(G(z_D))$ is reflected during gradient descent; $\lambda_k$ is a proportional gain for $k$ such as the learning rate in machine learning terms; and $\gamma \in [0,1]$ is a diversity ratio that results in high image diversity as it increases. Given that $k_t$ is changed in every training step to maintain $\mathbb{E}[\mathcal{L}(G(z))] = \gamma \mathbb{E}[\mathcal{L}(x)]$ for the equilibrium, global measure of convergence is regarded as the closest reconstruction $\mathcal{L}(x)$ with the minimum absolute value of proportional control algorithm error $|\gamma \mathcal{L}(x) - \mathcal{L}(G(z_G))|$ [31] as

$$M_{global} = \mathcal{L}(x) + |\gamma(\mathcal{L}(x)) - \mathcal{L}(G(z_G))| \qquad (18)$$

### 5.2. Proposed Generative Models

Although designers can utilize various state-of-the-art generative models in parallel for Stage 5 in the proposed framework (see Fig. 1), we chose and modified BEGAN architecture as illustrated in Fig. 3. The encoder of discriminator is a network consisting of five 3 × 3 convolutional layers, four 2 × 2 sub-sampling layers with stride 2, and one fully connected (FC) layer. For the dimension of each layer, w × h × n represents the width, height, and the number of kernels, respectively. The exponential linear unit (ELU) is used for the activation function. Generator and decoder of discriminator use a similar structure as this but by replacing sub-sampling to up-sampling. The model was trained with 16, 32, 64, and 128-dimensional latent variable **z**, and all the results were utilized. Adam optimizer was used with a learning rate of 0.00008, initial value of $k_t$ as 0, $\lambda_k$ as 0.001, $\gamma$ as 0.7, and minibatch size of 16 (see Section 5.1.2). The learning rate parameter is set with reference to the settings of previous papers that studied image quality by using BEGAN [48,49].

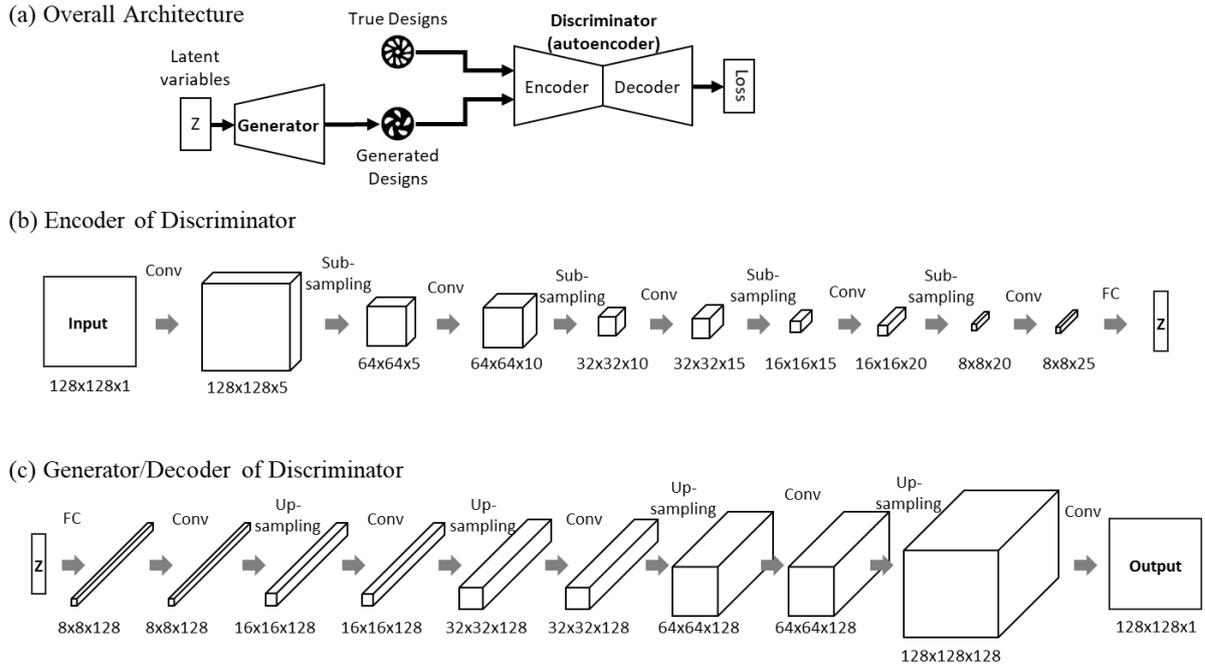

**Fig. 3.** Network architecture of BEGAN and autoencoder

In Stage 7, autoencoder is used to evaluate design novelty. Reconstruction errors, which are the loss functions of autoencoder, are widely used to detect anomaly [50]. The idea is that autoencoder can effectively reconstruct similar data to training data but not dissimilar data. This study assumes that design novelty can be measured the same way as anomaly detection. Previous designs in Stage 1 used training data for autoencoder, and then trained autoencoder can calculate reconstruction error of new designs generated by iterative design exploration. The new design which has the higher value of reconstructing error is regarded as the design which has the more novelty. This study employs the autoencoder structure which is used in BEGAN as the discriminator in Fig. 3, with the same hyper-parameter settings used. The equation of reconstruction errors is the same as Eq. (15).

In addition, a regression model can be used as an alternative to autoencoder when previous design data are insufficient. Topology design results (in Stage 2) at the first iteration have the similarity parameter ($\lambda$), so that regression model can be built where output $y$ is set to the similarity parameter. VGG16 [51], which is one popular CNN for regression model, was tested. Results show that CNN-based regression can also predict similarity which is contrary to novelty.

# 6. Results

This section shows the results of a case study applying the proposed framework to 2D wheel design. In Stage 1, frontal wheel designs were collected in the market by web crawling and converted it to binary images. A total of 658 binary images of the wheel are collected through post-processing as reference designs for the first iteration of iterative design exploration.

## 6.1. Iterative Design Exploration

### 6.1.1 Design Exploration by Topology Optimization (Stages 2 and 3)

Topology optimization is performed in parallel according to similarity and force ratio parameters listed in Table 1. Fig. 4 shows an example of optimized designs according to five levels of force ratio when the similarity is 0.0005. A large shear force is observed to make many thin spokes of whirlwind shape. On the other hand, large normal force makes thick and less curved spokes.

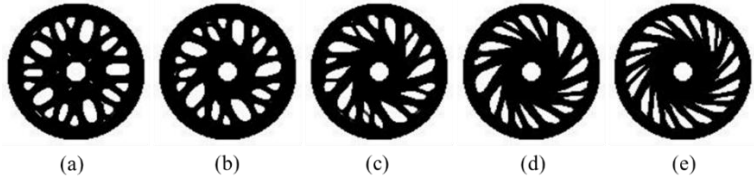

**Fig. 4.** Topology optimization of wheel design when the force ratio is set to (a) 0, (b) 0.1, (c) 0.2, (d) 0.3, (e) 0.4

Fig. 5 shows the optimal designs under five levels of similarity (i.e., (b) to (f)). (g) shows the reference designs, and (a) shows the result when reference design is unused. The ratio between normal and shear force is set to 0.1 in case of reference design A and 0.2 in case of reference design B, and the volume fraction is identical to reference designs. Table 2 lists the similarity and compliance of each optimal design. The optimal designs from topology optimization evidently indicate the tradeoff between engineering performance and similarity to the reference design.

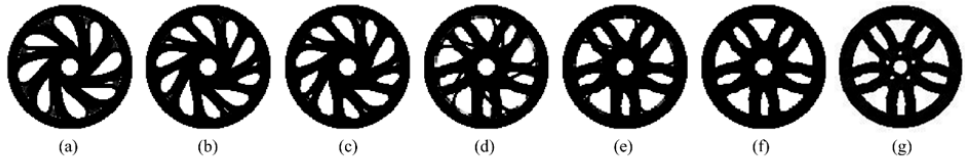

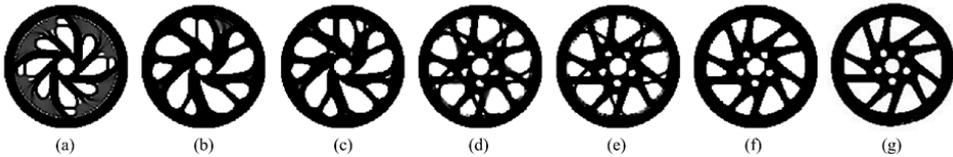

**Fig. 5.** Topology optimization results when the similarity is set to (a) 0, (b) 0.0005, (c) 0.005, (d) 0.05, (e) 0.5, (f) 5, and (g) reference design

**Table 2.** Similarity and compliance of each reference design in Fig. 5

| | Similarity | 0 | 0.0005 | 0.005 | 0.05 | 0.5 | 5 |
|---|---|---|---|---|---|---|---|
| Compliance | Reference design A | 5.28 | 5.23 | 6.17 | 7.14 | 8.87 | 9.28 |
| | Reference design B | 8.88 | 8.90 | 8.94 | 9.76 | 10.71 | 13.02 |

Fig. 6 demonstrates the effectiveness of the proposed objective function in Eq. (12), which states that the proposed method can yield different designs reflecting the shape of reference design when other boundary conditions are the same for all optimized design such as force ratio. The described results suggest that the proper range of similarity that optimal designs are continuously changed varies depending on the reference design. Therefore, an experimental investigation is encouraged in advance.

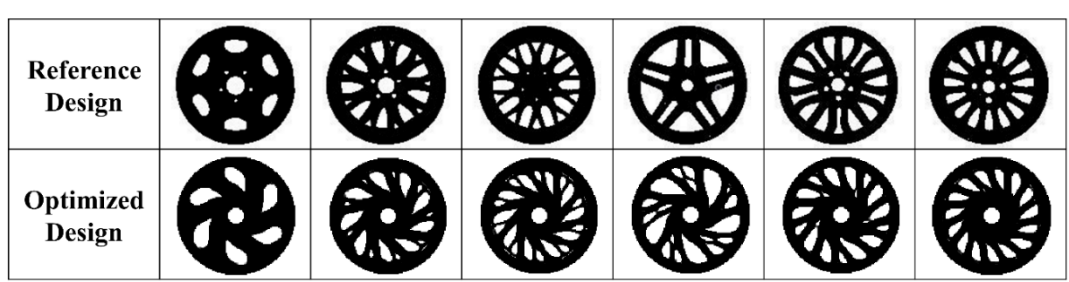

**Fig. 6.** Optimized designs under the same boundary conditions and different reference design

Consequently, 1619 new designs have been created after filtering at Stage 3. Stage 4 is passed automatically in the first iteration of the iterative design exploration, and the ratio of new designs is calculated as a criterion from the second iteration.

**6.1.2. Design Exploration by Generative Model (Stages 5 and 6)**

A total of 2277 designs were identified after Stage 4, where 658 previous designs were identified at Stage 1 and 1619 designs were obtained through Stage 4. These 2277 designs are used for training designs of BEGAN at Stage 5. The training takes around three hours on four GTX 1080 GPUs in parallel. Fig. 7 presents examples of 128 × 128 image from the BEGAN generator, and Fig. 8 shows that global convergence $\mathcal{M}_{global}$ is achieved without oscillation. A total of 128 designs were achieved through filtering at Stage 6. These designs were used as reference designs for Stage 2 at the second iteration of iterative design exploration. New 385 topology designs are regenerated after topology optimization (Stages 2 and 3).

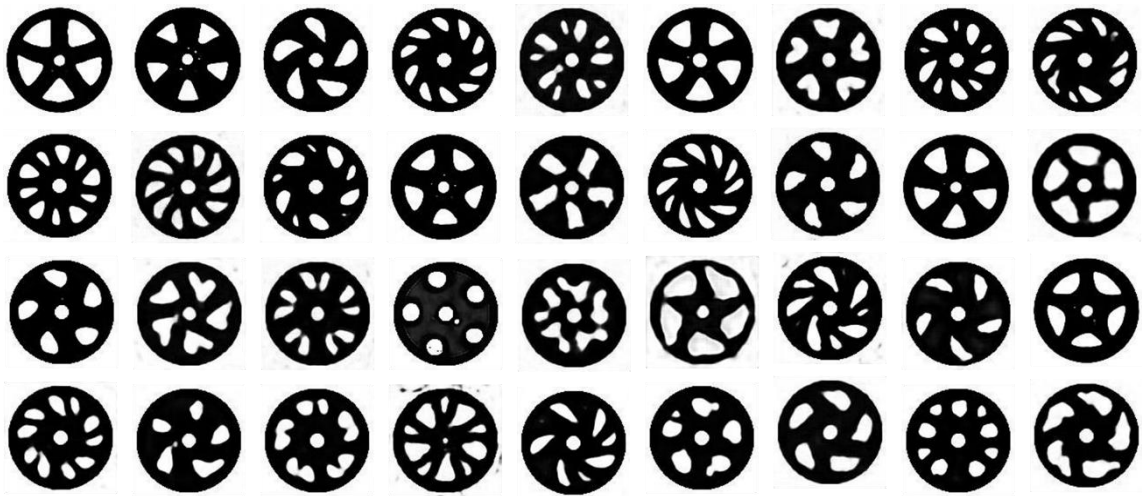

**Fig. 7.** Generated wheel designs by BEGAN

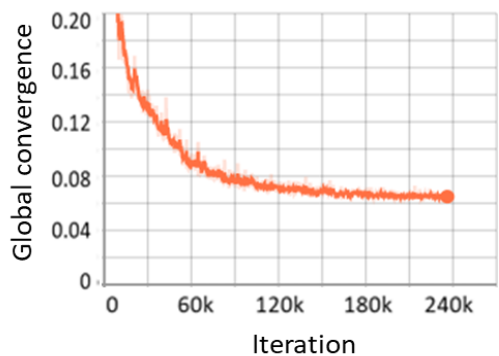

**Fig. 8.** Convergence results of BEGAN

As shown in Fig. 7, the BEGAN design results are roughly symmetrical, circular, and have holes in the center. Many GAN research studies in computer science use face dataset as benchmark data, and their results also demonstrate GANs capture symmetry features of human faces very well, without being taught [41,45].

**6.2. Design Evaluation**

**6.2.1. Novelty Evaluation by Autoencoder (Stage 7)**

For the autoencoder, 80% of the previous designs were used as training data, and 20% as test data. Fig. 9 shows examples comparing reconstruction results between test data of previous designs and generated designs. Designs similar to previous ones portray satisfactory reconstruction, while dissimilar designs portray otherwise.

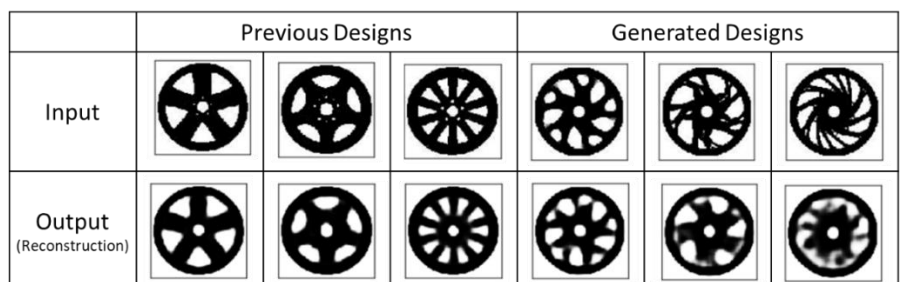

**Fig. 9.** Comparison between previous designs and generated designs in reconstruction of autoencoder

**6.2.2. Evaluation and Visualization (Stages 8 and 9)**

Finally, 2004 new designs are generated after two iterations, which are not included in the previous design set. Table 3 summarizes the number of input and output designs used at each stage. The termination criteria calculated at Stage 4 after two iterations is 23.8% (385/1619 = 23.8) which is less than the threshold of 0.3.

**Table 3.** Number of new generated designs at each stage

| Iteration | First | | Second |
|---|---|---|---|
| Method | Topology Optimization (Stages 2 & 3) | BEGAN (Stages 5 & 6) | Topology Optimization (Stages 2 & 3) |
| Input | 658 | 2277 (= 1619 + 658) | 128 |
| Output | 1619 | 128 | 385 |
| New topology designs (accumulated) | **1619** | - | **2004** (= 1619 + 385) |

Examples of design options are shown in Fig. 10. A 3D scatter plot for 2004 design options was crafted by using three design attributes (novelty, cost, and compliance) as an axis (Fig. 11(a)). Each attribute value used in the plot is normalized from 0 to 1. Fig. 11(b) shows that trade-offs between compliance and cost make a smooth Pareto curve, because designs are all topologically optimized. On the Pareto curve, two designs are shown as examples, one with the lowest cost and the other with the lowest compliance. Figures 11(c) and 11(d) show trade-offs between novelty and other attributes. Assuming novelty is a positive trait, two designs located on the Pareto curves are shown as examples in the figures, respectively.

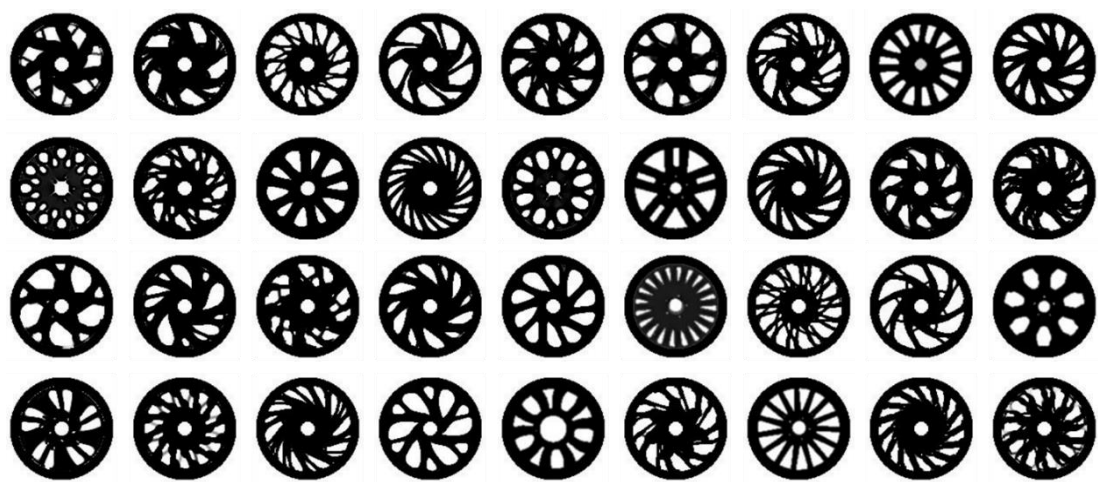

**Fig. 10.** Generated design options

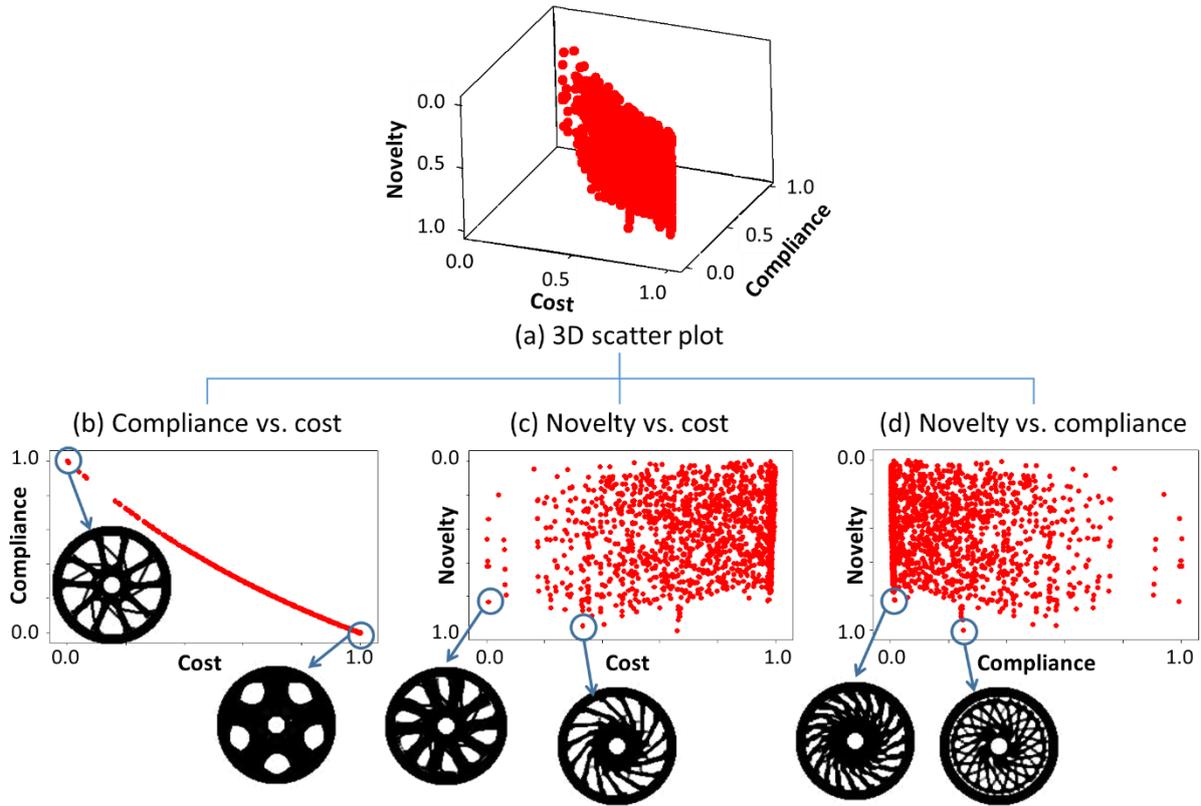

**Fig. 11.** Visualized design options by three attributes: novelty, cost, and compliance

Designers can trade-off three attributes, and select designs according to their design purpose and preference. Then, they can create a 3D design based on a 2D design and prototype it by 3D printing. Fig. 12 shows an example of a 3D wheel design (i.e., STL file for 3D printing) after selecting a 2D design.

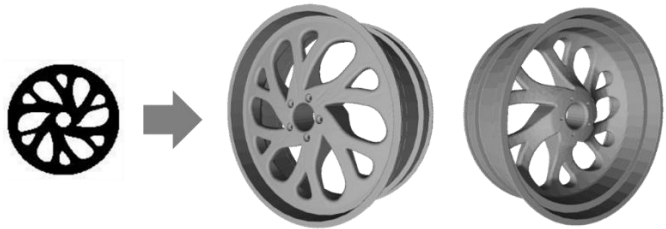

**Fig. 12.** An example of a 3D wheel design using the selected 2D design

## 7. Discussion

This section analyzes and discusses performance and necessity of the main methods used in the proposed framework.

### 7.1. Topology Optimization

As an additional analysis, to check necessity of the reference designs, we conducted topology optimization without reference designs as shown in Fig. 13. Results without the reference designs as the benchmark would converge to an identical optimum if there is no change on boundary conditions. So it cannot yield aesthetical diversity. Also, it sometimes fails to converge, when it starts from uniform density and has no shear force since the displacement caused by the normal force exerted on the surface is almost uniform (see Fig 13(a), (b)). Therefore, the reference designs can enhance the diversity of designs while achieving convergence.

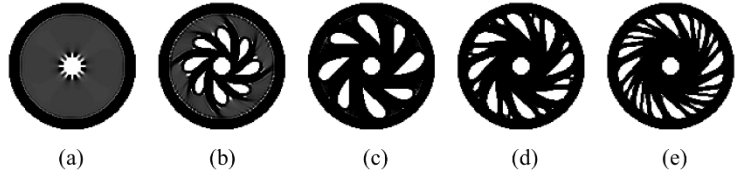

**Fig. 13.** Topology optimization without reference designs when the force ratio is set to (a) 0, (b) 0.1, (c) 0.2, (d) 0.3, (e) 0.4

In addition, topology optimization can theoretically generate infinite designs without the help from generative models, when topology optimization results are used as reference designs for topology optimization in the next iteration. We used the topology optimization results of Fig. 4 (b), (c), (d), and (e) as reference designs for the next topology optimization. Fig. 14 shows the selected results which were most different from the reference designs. The results show that iterative topology optimization generates similar designs since their original parent (reference design) is the same. This can be an empirical evidence that reference designs with fundamentally different topologies are needed to obtain a diversity of generated designs, and generative models make this possible. Therefore, we do not use topology optimization results as reference designs in the proposed process.

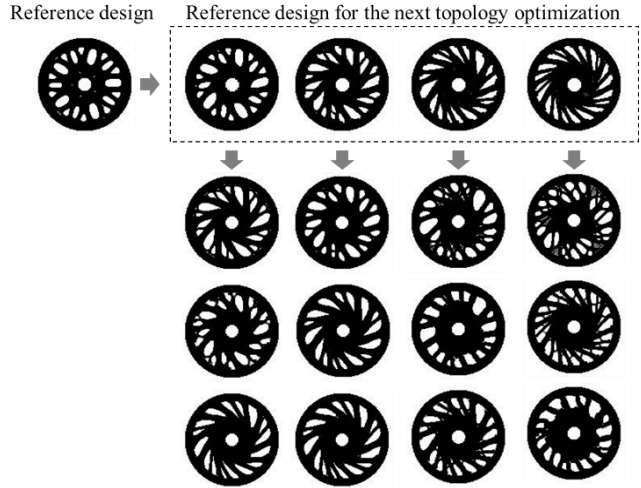

**Fig. 14.** Iterative design exploration by topology optimization only

## 7.2. BEGAN

One of the main problems of GAN is that there is no standardized method of measuring model performance. The following are some ways. First, one can check in person if generated data looks reasonable. Second, convergence criteria can be checked as shown in Fig. 8. Third, we can compute the difference between real data and generated data in feature space, for example, by using Inception Score (IS) or Fréchet Inception Distance (FID) [52]. In our proposed method, we take the second approach and check the global convergence for model validation. One advantage of our framework is that generated designs by BEGAN are not used as final output but as input for topology optimization. Therefore, even though the performance of BEGAN is slightly low, the framework can still work robustly.

Other than BEGAN, we additionally tested other generative models such as DCGAN and VAE which were introduced in Section 5.1. In our experiments, DCGAN and VAE display similar designs as shown in Fig. 15. They appear to produce more detailed and complex shapes than BEGAN, but they are blurrier and less symmetrical. In addition, the generated designs lack novelty in that the designs are more similar to those in the train data. We also tested these results as reference designs for topology optimization, but the results produced are less diverse.

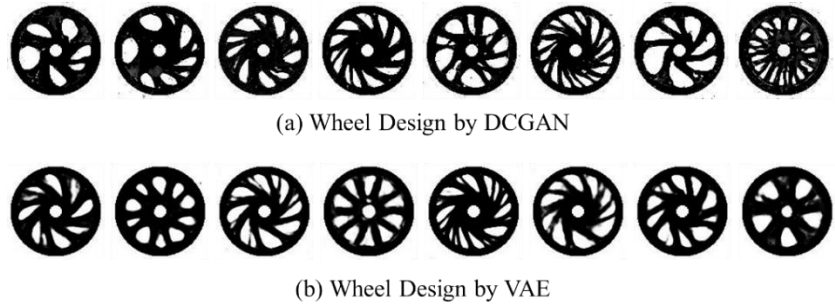

**Fig. 15.** Example of generated designs by DCGAN and VAE

In sum, we acquire some empirical insights for utilizing generative models in design exploration. First, BEGAN is a good choice for generating reference designs, because it tends to create topologically novel, yet simple, designs. Detailed shapes of reference designs used in topology optimization fail to generate diverse designs. In fact, CAE softwares also have a feature that simplifies CAD models before conducting topology optimization (e.g., filling up holes). In future research, we plan to train a network that selects only the "good" reference designs from the results of multiple generative models. This is because DCGAN and VAE could also have some reference designs that BEGAN cannot create. Second, designers have to try different latent space dimensions and epochs. In our study, we varied dimensions for latent variables (i.e., 16, 32, 64, and 128), saved the models at different epochs, and ultimately obtained data from many variations of the model. It is not practical to select only one latent dimension or epochs because each model generates unique designs.

### 7.3. Autoencoder

To validate the performance of autoencoder more quantitatively, we tested how much the model can classify previous designs and generated designs, assuming generated designs have more novelty than previous designs. We select 131 test data set (20% of 658 previous designs) for previous design and generated design, respectively, and create a confusion matrix as shown in Fig. 16. The autoencoder calculates a reconstruction error for 262 test designs, and we sort them by error size. We classify the top 50% of the designs as generated designs, and the bottom 50% as previous designs. In FP and FN cases which represent the designs that were misclassified, we see that it is not easy to distinguish between previous and generated designs even by human eye. As measuring criteria, both precision and recall are 91.6%.

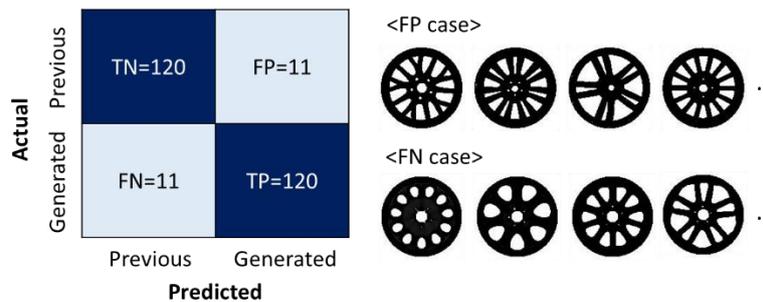

**Fig. 16.** Confusion matrix for autoencoder

## 8. Conclusion

This study proposes a design automation framework that generates various designs ensuring engineering performance and aesthetics, and its effectiveness is demonstrated by 2D wheel design case study. The contribution of this research can be addressed as follows:

First, this research considers engineering performance and aesthetics simultaneously. The proposed framework is able to control the similarity with reference designs and engineering performance as a multi-objective function.

Second, a large number of designs starting from a small number of designs was generated. An iterative process is proposed where topology optimization is conducted to create training data for the generative models, and output designs from generative models are used as reference designs for topology optimization again.

Third, the proposed framework offered diverse designs in comparison with the conventional generative design. Moreover, increased diversity is accounted to the use of reference designs generated by generative models.

Fourth, the robustness on quality of designs is improved. The conventional generative models are prone to induce the mode collapse and large variance of the quality. However, results of the generative model in the proposed framework are refined through topology optimization instead of direct utilization (e.g., post-processing).

Finally, a comparison between the novelty of generated design and the previous designs can be evaluated. The reconstruction error of autoencoder is used for the index of similarity to existing designs.

This research is performed on a 2D design space and pixel-wise images, which is identified as its main drawback. Thus, a 3D design application with voxel data should be further investigated for practical design, and various case studies should be tested. In addition, a recommendation system that suggests the appropriate designs (i.e., predicting the preference of designers and consumers among design candidates) will be carried out.

# Acknowledgement

This work was supported by the National Research Foundation of Korea (NRF) under Grant (No. 2017R1C1B2005266) and NRF under Grant (No. 2018R1A5A7025409). The authors would like to thank Ah Hyeon Jin, Seo Hui Joung, Gyuwon Lee, and Yun Ha Park who are undergraduate interns of Sookmyung Women's University, for pre- and post-processing of data, and also Yonggyun Yu of Korea Advanced Atomic Research Institute for his advice and ideas.